\documentclass[a4]{article}

\usepackage{ijcai-derived}

\usepackage{times}
\usepackage{soul}
\usepackage{url}
\usepackage[hidelinks]{hyperref}
\usepackage[utf8]{inputenc}
\usepackage[small]{caption}
\usepackage{subcaption}
\usepackage{graphicx}
\usepackage{amsmath}
\usepackage{amsthm}
\usepackage{booktabs}
\usepackage{algorithm}
\usepackage{algorithmic}
\usepackage[switch]{lineno}

\title{Multi-Agent Systems in Emergency Departments: Validation Study on a ED Digital Twin}

\author{
Markus Wenzel$^{1,2}$
\and
Tobias Strapatsas$^3$\and
Jessika Kress$^3$\and
Dorothea Sauer$^3$\and
Nele Gessler$^3$\And
Horst K. Hahn$^2$\\
\affiliations
$^1$Constructor University, Bremen, Germany\\
$^2$Fraunhofer Institute for Digital Medicine MEVIS, Bremen, Germany\\
$^3$Asklepios Kliniken Hamburg GmbH, Hamburg, Germany\\
\emails
mwenzel@constructor.university,
\{t.strapatsas, j.kress, do.sauer, n.gessler\}@asklepios.com,
horst.hahn@mevis.fraunhofer.de
}

\begin{document}

\maketitle

\begin{abstract}
    Emergency departments (ED) face challenges in patient care and resource management. We propose to explore optimization strategies in a realistic and flexible model and develop a hybrid Discrete Event Simulation (DES) and Agent-Based Model (ABM) simulating highly configurable ED environments. We specifically focus on the validation of the modeling approach. We derive configurations for ED sizes, patient load, and staffing from real-world studies. We then validate the model expressivity by matching its key performance indicators and metrics with their values known from literature. We proceed by implementing scientifically established and practice-proven resource optimization strategies. Comparing the documented real-world outcomes with our model's results demonstrates that the DES-ABM based simulation can effectively replicate real-world ER dynamics under interventions. We lastly integrate a Proof-of-Concept multi-agent system (MAS) that can autonomously explore resource allocation strategies within the simulated ER environment based on a temporal ledger of ED event records. This modular DES-ABM-MAS framework offers a powerful tool to explore resource optimization strategies in emergency departments.
\end{abstract}

\section{Introduction}
Emergency departments (EDs) are under pressure due to rising patient volumes, resource constraints, and the need for efficient patient flow management. To understand and research these challenges, simulation models have become invaluable tools. They can reproduce ED dynamics to help to evaluate potential interventions. We propose to use a hybrid modeling approach combining Discrete Event Simulation (DES) and Agent-Based Model (ABM) of fictional EDs of different sizes. Our main objective is to validate the simulation model against scenarios with realistic settings and to reproduce resource optimization strategies aimed at enhancing ER efficiency.

We conduct the validation by comparing key performance indicators (KPIs) recorded from the simulation with empirical data reported in scientific studies of EDs to ensure the model's accuracy and reliability. The validated model serves as a platform to investigate various interventions documented in literature and evaluated in practice, such as staff allocation adjustments and patient flow management techniques. By simulating these strategies applied to baselines we constructed to exhibit specific bottlenecks, we aim to provide a provably realistic simulation of baselines and interventions in an open-source, Python-based modular and flexible simulation environment. Our quantitative results demonstrate that the DES-ABM can effectively replicate real-world ER dynamics, including interventions into patient flow and staffing. 

A prototypical integration of a LLM (Large Language Model)-based Multi-Agent System interacting with the simulation provides proof of concept of its modular extensibility to research state-of-the-art AI (Artificial Intelligence) approaches for ED optimization.

\section{State of the Art}
\label{sec:state_of_art}
Emergency departments (EDs) have been the focus of extensive research due to their critical role in healthcare systems and the challenges they face. Various modeling approaches have been employed to study ED operations, with Discrete Event Simulation (DES) and Agent-Based Modeling (ABM) being among the most prominent. DES has been widely used to simulate patient flow and resource allocation in EDs, allowing researchers to evaluate different operational strategies and their impact on KPIs such as wait times, length of stay (LoS), and patient satisfaction \cite{Morgan2011,Hajrizi2019}. ABM, on the other hand, provides a more granular approach by modeling individual behaviors and interactions, enabling the study of complex dynamics and emergent phenomena within the ED environment \cite{north2007managing,macal2010tutorial}.

Efforts exist to model ED scenarios using different approaches, where in the review papers a clear dominance of purely DES based approaches is visible \cite{salmon2018structured}. While powerful, \cite{salmon2018structured} implement their model in Java integrated into AnyLogic, whereas our implementation is custom-built in pure Python to allow for greater flexibility and integration with AI-based optimization modules in the future. \cite{ghanes2015simulation} is among the more recent studies using DES for simulation-based optimization of staffing levels. \cite{liu2017agent}, as one exception, present an ABM approach to quantitatively analyze and predict behaviors in EDs, focussing on patients and staff. 

\cite{terning2022modeling} shows a hybrid DES-ABM approach technically similar to our work. A DES layer manages resources, but agents follow pathways in a DES-like approach. Their focus is on pandemic scenarios, where patient arrivals and pathways differ from standard operations. In contrast, our model incorporates detailed agent behaviors and interactions, models resources as agents, and only the general disease treatment pathways for different diseases using DES. Likewise, \cite{Liu04032025} use a hybrid DES-ABM approach to explore resource scheduling strategies combined with queue management policies and evaluate the impact on stay times for discharged patients as well as those who are hospitalized. Their focus is on optimizing scheduling and queue policies, whereas our work emphasizes validating the simulation model against real-world data and exploring a broader range of proven and novel interventions.

Studies have explored various interventions aimed at improving ED efficiency. For instance, Fast Track systems have been implemented to expedite the treatment of low-acuity patients, resulting in reduced wait times and improved patient throughput \cite{Bonalumi2017,Berger-vonOrelli2025}. Split-Flow models, which involve segregating patients based on acuity levels and directing them to different care pathways, have also shown promise in enhancing ED performance by optimizing resource utilization and minimizing bottlenecks \cite{hwang2012solutions,bish2016ready}. Additionally, staffing strategies, such as adjusting nurse-to-patient ratios and implementing flexible scheduling, have been investigated for their potential to alleviate overcrowding and improve patient outcomes \cite{Griffiths2023,Wang2023}.

Challenges remain. Many studies have highlighted the importance of incorporating empirical data into simulation models to enhance their validity and reliability \cite{Centeno2013,Ouda2025}. Furthermore, the integration of DES and ABM approaches has been proposed as a means to leverage the strengths of both methodologies, providing a more comprehensive understanding of ED operations and facilitating the evaluation of multifaceted interventions \cite{north2007managing,macal2010tutorial}.

Despite the wealth of research in this area, there is a lack of generalizing models that are not tailored to one specific ED or use case. We thus propose a combined DES-ABM of an emergency department implemented in pure Python. Our primary objective in this contribution is to validate the simulation model against real-world data to prove its utility to explore resource optimization strategies. The model aims to serve as a platform to investigate and understand interventions documented in literature and evaluated in practice, but more importantly to explore strategies incorporating machine learning methods learning from the simulation data in our "ED Digital Twin". With that in hand, we aim to provide an environment to explore the practical utility of LLM-based multi-agent systems in such a setting, thereby pushing the state of the art in ED simulation and optimization beyond exploratory studies like \cite{Han2024LLM_ED} and others reviewed in \cite{Preiksaitis2024}.

\section{Simulation Methods}
\label{sec:methods}

\subsection{Architecture Overview}

We implement a {hybrid Agent-Based Modeling (ABM) and Discrete Event Simulation (DES)} architecture which uses the Mesa ABM Framework \cite{mesa} to manage agent behaviors, spatial interactions, and step-based simulation dynamics and the SimPy DES Engine \cite{simpy} to handle resource allocation, queuing, and event scheduling for patient treatment processes. This hybrid approach allows for both emergent behavioral dynamics (ABM) and precise resource-constrained event timing (DES) \cite{gul2015comprehensive,Kar2025}.

\begin{figure*}[t]
    \centering
    \includegraphics[width=.9\linewidth]{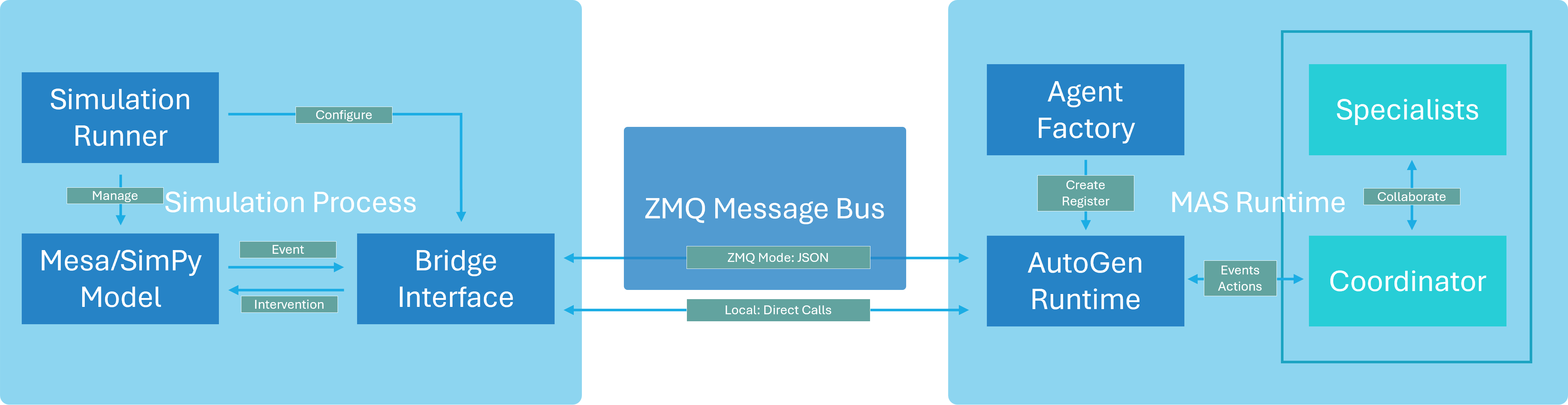}
    \caption{Hybrid ABM-DES Simulation Architecture Overview}
    \label{fig:architecture_overview}
\end{figure*}

\subsubsection{Agent-Based Modeling Layer}

The ABM layer treats everything that bears spatial references, can move about, or acts based on an individual history as agents. It is important to note that ABM agents are not stateless; instead, they can even be equipped with memory and experiences -- they live in time. 

Among the modeled agents are the \texttt{PatientAgent} that represents patients in the ED who follow treatment pathways, queue for resources, and potentially deteriorate in disease severity. \texttt{DoctorAgent}s provide treatments based on specialization. They accumulate fatigue, and in consequence may cause errors. \texttt{NursePractitionerAgent}s are mid-level providers (NPs/PAs) who staff Fast Track units and treat low-acuity patients (ESI levels 4-5). To support the staff and physical resources, \texttt{AssistantAgent}s like nurses and techs assist with treatments. They also experiences fatigue.

Rooms and equipment are modeled as \texttt{RoomAgent} and \texttt{EquipmentAgent}, respectively, and have capacity constraints and availability restrictions.

All agents live in a spatial environment giving the ED layout, with a waiting area, treatment rooms, diagnostic facilities, and staff areas in a grid-based spatial model, where agents occupy grid cells. Multiple agents can occupy the same cell, enabling realistic modeling of crowded waiting rooms or shared treatment spaces. In our simulation, movement and agent counts are constrained by the room layouts, i.e. agents cannot pass through walls or overcrowd small rooms. Staff-patient interactions and patient treatment require spatial proximity between the agents involved.

\subsubsection{Discrete Event Simulation Layer}

Resource management is handled by the DES layer using SimPy. Key components of the SimPy resources model include \texttt{PriorityResource}s like treatment rooms, for which higher severity patients get priority. Equipment and staff time are modeled as general \texttt{Resource}s. To handle complex resource requirements (e.g., a procedure needing both a room and specific equipment), the simulation uses \texttt{AllOf/AnyOf} Compound resource requirements to group multiple resources together.

We model certain event types impacting the DES layer. Those include Patient arrivals modeled stochastically based on configurable rates; Treatment events with variable durations affected by procedure type and staff fatigue; resource requests which are queue-based with priority scheduling; and patient state transitions reflecting pathway progression and potential deterioration.

\subsubsection{System Architecture}

The system is designed to be modular and flexible. Figure \ref{fig:architecture_overview} provides an overview of the architecture, showing the interaction between the ABM and DES layers, key agent types, resource management, and data collection components. It also shows a tentative interface to a Multi-Agent System based AI optimization module that we have implemented as a proof-of-concept but did not further evaluate in this study. Further modulaer implementations comprise batch execution modules (used in the validation study) and a web frontend used to observe the temporal evolution of the simulation visually.

\subsection{Simulation Aspects}

\paragraph{Patient Arrival Modeling.}

Patient arrivals are modeled using a {Non-Homogeneous Poisson Process (NHPP)} to reflect realistic daily and weekly variability in emergency department demand \cite{channouf2007application}. The daily arrival rate $\lambda(t)$ follows a piecewise linear shape derived from standard ED benchmarking data \cite{wiler2015emergency}. Weekly patterns are additionally applied as multiplicative factors to the daily rate. The instantaneous arrival rate 
is from those patterns calculated 
%
%
based on the target average arrivals per hour, a daily shape multiplier at hour $h$, and a weekly seasonality multiplier for day $d$.

To ensure consistent patient populations across different experimental scenarios (e.g., comparing staffing interventions), the simulation uses a dedicated random number generator for patient generation. This isolates the patient arrival stream (timing, acuity, condition) from the internal simulation dynamics (e.g., staff fatigue checks, mortality events), ensuring that interventions are tested against identical patient loads for meaningful validation studies.

\paragraph {Patient Severity \& Triage Modeling.}

The simulation distinguishes between true acuity based on the Emergency Severity Index (ESI) 1-5 scale and the assigned triage level to model real-world uncertainty and errors. Patients are generated with a true acuity distribution reflecting a typical ED case mix with a focus on mid-acuity cases \cite{wiler2015emergency}. 

Triage nurses assign a priority level based on the patient's condition. This process is modeled probabilistically to include realistic error rates (Under/Over-triage). The simulation uses the nurse-assigned triage level for resource prioritization and routing, but uses the true acuity (ESI) level for deterioration and mortality calculations. The ESI level distribution is not adapted to different ED sizes in our study but could be in future work.

\paragraph{Pathway-Driven Patient Journeys.}

Patients follow {configurable pathways} defined in YAML. Each pathway specifies a sequence of treatment steps, base durations, and initial severity. Note that infectious diseases or pandemic situations are not specifically regarded; there are also no isolation rooms defined yet, impacting staff allocation and durations. 


Pathways are DES components with {deterministic ordering}. Their steps are executed in the defined sequence, though some pathways probabilistically branch based on patient response or test results. {Duration variability} modifies base times by staff fatigue and patient factors, linking DES and ABM layers. Patients arrive untriaged with a given true ESI; nurses assign severity levels (1-5) based on presenting conditions.  Patients are assigned to doctors with appropriate specializations (e.g., trauma, general) and specific equipment, leading to realistic resource competition and queuing.

\paragraph{Shift Scheduling Models.}

We model staff availability and fatigue dynamics by implementing distinct scheduling strategies for different roles, reflecting common hospital practices \cite{rogers2004working}. Nurses follow a block model with 12-hour shifts to minimize handovers and ensure continuity of care. Doctor scheduling is configured according to a waterfall model, where they work shorter, higher-intensity shifts with staggered start times to match patient arrival patterns.

\paragraph{Fatigueing, error probability, and slowdown model.}

The fatigue model is grounded in clinical research relating fatiguing to medical errors and sleep deprivation \cite{west2009association,durmer2005neurocognitive}. Fatigue accumulates linearly during active work. Recovery occurs during rest periods, which the agents try to take regularly depending on workload. This can lead to realistic over-fatigue in high-demand situations.

Based on \cite{west2009association}, error probability follows an {exponential relationship} with fatigue, yielding about a 5.5\% error probability at full fatigue.

The simulation models the impact of fatigue and shift duration on staff performance through two mechanisms: cognitive effectiveness (shift duration), and fatigue-based slowdown. \cite{james2020sleep} showed that cognitive function remains relatively stable for the first 8 hours of a shift but degrades significantly thereafter.

According to \cite{durmer2005neurocognitive}, fatigue also causes linear performance degradation, ultimately slowing down task execution. A maximum slowdown factor defines the limit of this effect. Fatigue leads to increased error probability during treatments, which can worsen patient severity. Default error rates are set to reflect typical medical error rates \cite{west2009association}. Fatigue parameters are configurable per staff role.

Treatment durations are dynamically scaled based on staff fatigue and cognitive effectiveness. This mechanism ensures that as staff become fatigued, process times increase, contributing to realistic bottlenecks and LoS variability, ultimately resulting in increased fatality rates if staff is over-fatigued.

\paragraph{Mortality Model.}

The mortality model is based on clinical research on emergency department outcomes, particularly the relationship between severity, wait time, and patient mortality \cite{nicholl2007relationship,guttmann2011association,sun2013effect}. We calculate a per-step mortality probability 
%
%
from the base mortality rate by {true acuity (ESI)} level, an additional risk per hour waiting, and multipliers increasing if treatment errors occurred or staffing ratios deteriorate \cite{aiken2002hospital}. Based on \cite{nicholl2007relationship}, baseline mortality rates vary by patient severity, which we configured accordingly.


Beyond this, we double the mortality probability after treatment errors that worsen patient severity. Also patient-to-nurse ratios $>$ 4:1 increase mortality risk by 7\% per additional patient \cite{aiken2002hospital}.

\subsection{Intervention Modeling}

For validation and experimentation, the simulation supports dynamic intervention modeling. Interventions can modify pathways, resource allocations, or staff behaviors at runtime to simulate process changes. Three interventions are implemented, based on literature evidence \cite{rowe2011use,Bonalumi2017,spetz2004california}.

\paragraph{Split Flow / Provider in Triage (PIT).}

The simulation supports a "Split Flow" or "Provider in Triage" (PIT) intervention, designed to improve throughput for mid-acuity patients (typically ESI 3) \cite{rowe2011use}. PIT places a provider (Doctor or Advanced Practice Provider) at the triage point to order labs and imaging before the patient is assigned a main bed ("front-loading"), treat and discharge low-complexity patients directly from triage, and identify patients who don't have to be treated in beds.

The intervention is implemented as a {dynamic pathway transformation}, which inserts a step immediately after triage for eligible patients. Any subsequent diagnostic steps (Imaging, Labs) are moved \textit{before} the main exam room assignment. This allows diagnostics to be completed earlier, reducing overall length of stay. This intervention requires a dedicated triage doctor resource to avoid conflicts with main ED operations.

\paragraph{Nurse Ratio.}

The Nurse Ratio intervention enforces a maximum patient-to-nurse ratio (e.g., 4:1) by dynamically adjusting nurse staffing levels based on current patient load \cite{spetz2004california}. If the ratio exceeds the threshold, additional nurse resources are activated (if available) or patient admissions are temporarily paused until ratios normalize. This intervention aims to reduce nurse fatigue and improve patient care quality.

\paragraph{Fast Track.}

The Fast Track intervention dedicates a treatment area to low-acuity patients. It is staffed by Nurse Practitioners (NPs) or Physician Assistants (PAs) \cite{Bonalumi2017,Wu2021}. These ESI level 4-5 patients are routed to the Fast Track unit immediately after triage, bypassing the main ED flow. This reduces congestion in the main treatment areas and allows for quicker turnaround of minor cases. A special fast track room has to be dedicated for this purpose, along with a separate NP/PA resource pool.

%
%
%
%

\subsection{Process Metrics \& Validation}

To ensure simulation fidelity, we implemented baseline simulations of differently sized EDs with realistic staffing and patient arrival rates where key process metrics are tracked and validated against national benchmarks \cite{cdc2021nhamcs}. Our ED sizes range between 4-16 doctors and 2-32 nurses, with week-averaged arrival rates between 4-20 patients per hour (Table \ref{tab:er_sizes}). The floor plans are designed accordingly (compare Appendix, \ref{fig:floorplans}).

\begin{table}[th]
\centering
\caption{Emergency Room Resource and Staffing Configurations. Imaging rooms are subdivided into X-ray, CT, and Ultrasound rooms.}
\label{tab:er_sizes}
\begin{tabular}{llccc}
\hline
    &   \textbf{Parameter} & \textbf{M} & \textbf{L} & \textbf{XL} \\ \hline
\textit{Base Staffing} & Doctors & 4 & 10 & 16 \\
    &   Nurses & 2 & 5 & 32 \\
\textit{Patient Load} &  Avg. Arrivals/hour & 4 & 8 & 20 \\
    &   Approx. Yearly  & 35k & 87k & 175k \\
\textit{Room Counts} &  Triage  & 1 & 2 & 4 \\
    &   Exam  & 4 & 10 & 20 \\
    &   Shock  & 1 & 2 & 4 \\
    &   Imaging  & 1 & 2 & 4 \\ \hline
\end{tabular}
\end{table}

To help experts identify system constraints and optimize resource allocation, the simulation tracks queue lengths to perform bottleneck analysis. We record the queues lengths for every resource (rooms, staff, equipment) at each simulation step. Wait times are also accumulated per resource type to pinpoint bottlenecks. This helps detecting the most constrained resource based on queue-to-capacity ratios and utilization rates. These metrics are purposefully not forwarded to the AI optimization module (see Section \ref{sec:mas_integration}) to acknowledge that they are not easily observable in a real-world ED and thus unlikely available to an AI.

%
%
%
%

%
%

\subsection{Validation Approach}

The validation study assesses the impact of the three implemented interventions (Split-Flow, Nurse Ratio, Fast Track) across three different ED sizes (cf. Table \ref{tab:er_sizes}) to ensure robustness and scalability of the findings. Each combination of ED size and intervention is simulated for 30 runs (3 simulated days each; once baseline, once intervention) to assess statistical significance, resulting in more than 550.000 simulated ED patients across 540 simulation runs.

\paragraph{Validation Methodology.}

To rigorously evaluate the impact of specific interventions, the simulation employs a {targeted baseline approach}. Rather than comparing all interventions against a single generic baseline, each intervention is tested against a specific baseline configuration designed to exhibit the particular bottleneck that the intervention addresses. The Fast Track intervention is tested under high arrival rates (1.5x) to assess its ability to reduce LoS for ESI 4-5 patients by shielding them from high-acuity competition \cite{Berger-vonOrelli2025}; the Nurse-to-Patient Ratio intervention is evaluated in a ``Stressed Staffing'' scenario to measure the trade-off between throughput constraints and reduced staff fatigue \cite{aiken2002hospital}; and the Split-Flow intervention is applied to high-arrival scenarios to determine its effectiveness in reducing triage bottlenecks through early physician assessment \cite{rowe2011use}.

To ensure that interventions are actively influencing the simulation dynamics, specific counters are tracked:

\begin{description}
    \item [Fast Track Count] Tracks the number of low-acuity patients (ESI 4-5) explicitly routed to Fast Track resources.
    \item [Nurse Ratio Blocked Count] Tracks the number of simulation steps where a patient was ready for an exam room but was blocked specifically because the nurse-to-patient ratio limit was reached (even if rooms were available).
    \item [Physician Triage Count] Tracks the number of patients who received a physician evaluation during the triage phase.
\end{description}

This approach ensures that validation results reflect the {efficacy} of the intervention under the conditions it was designed for, avoiding false negatives where an intervention appears ineffective simply because the relevant bottleneck was not present in the baseline.


\subsection{Multi-Agent System Integration}\label{sec:mas_integration}

The simulation integrates a {Multi-Agent System (MAS)} based on LLMs (Large Language Models) as a cognitive layer to introduce experimental intelligent decision-making and resource optimization. This integration follows a "Digital Twin" architecture. The MAS observes the simulation state and intervenes to optimize outcomes. We employ a {Blackboard Architecture} where multiple specialized agents (Data Analyst, Staff Scheduler, Ressource Manager, Patient Flow Manager) observe the simulation state (the "Board") and sequentially propose actions from their perspectives. This happens in a batch-wise scheme. While the simulation is suspended after one batch (equalling a set number of simulation steps), each specialist receives a custom tabular view of the logged data of the preceding interval, emulating a real-world situation where those agents are tied to specific sensors and tailored to argue about their history. An Orchestrator agent receives and synthesizes all their proposals into actions it posts to the Board. This decouples the simulation from the specific agent implementations and allows for integration of different decision aids, including risk models or, like in our case, an LLM-based multi-agent system.

The MAS is implemented in a separate, decoupled thread based on Microsoft AutoGen \cite{autogen2024}, leveraging its event-driven architecture to manage agent communication and message routing. We employ a locally running GPT-OSS-20b model (via LM Studio \cite{lmstudio2024}) to power the agents, using a custom prompt design to elicit reasoning and structured proposals. The MAS architecture is designed to be modular, allowing for easy substitution of different LLM backends (e.g., Azure OpenAI, local models) via an Agent Factory pattern.

Using this setup, the MAS Orchestrator is asked to propose interventions. For this to work, the simulation runs in batches of a given number of steps. After each batch, the simulation event data of this batch is archived as a baseline trajectory. The MAS may suggest interventions at a timepoint $t_i$ within the timeframe of the last batch. Then, the system replays the simulation from the last checkpoint up to timepoint $t_i$ and any MAS interventions proposed to be injected are applied. The scenarion then continues until the batch finishes. This batched stepping allows the MAS to act with "hindsight" into one batch (but currently not into historic data or own memory). The injected intervention leads to a new trajectory from time $t_i$ and produces paired datasets (Baseline vs. Intervened) for reinforcement learning or to extract knowledge about successful interventions. Note that more than one intervention at different timepoints can be suggested for one batch, but they would not lead to as many new trajectories but all injected into the same one.

\section{Results}
\label{sec:validation_results}

\subsection{Validation Study Quantitative Results}

\begin{figure*}[t]
    \centering
    \includegraphics[width=\linewidth]{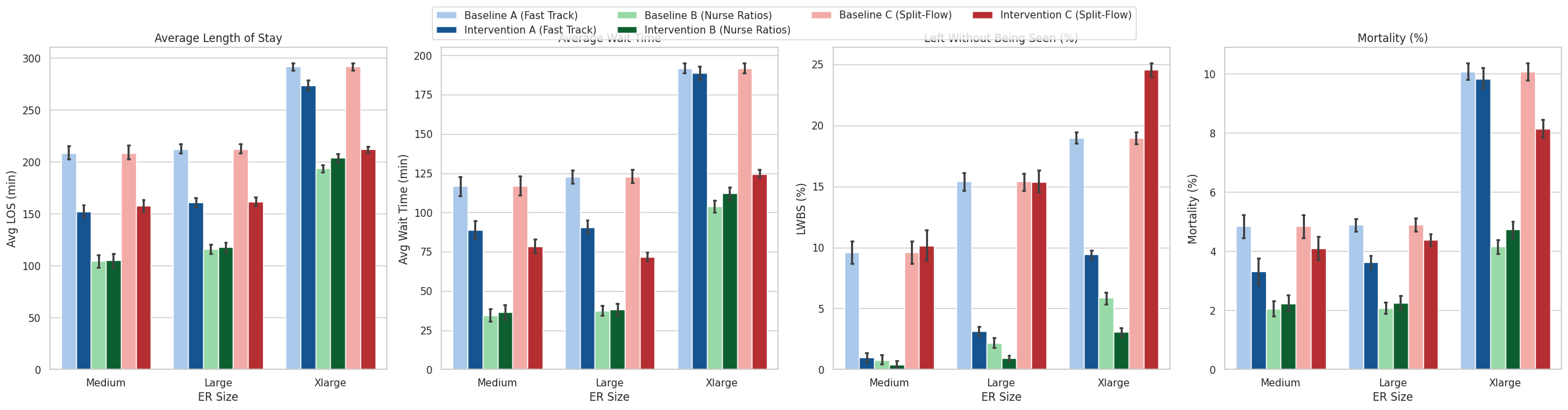}
    \caption{Analysis of performance metricss across ED sizes and interventions. Like colors (blue, green, red) represent the interventions within each size category, and their shades baseline vs. intervention outcome.}
    \label{fig:unified_metrics}
\end{figure*}

In Figure \ref{fig:unified_metrics} we show the average performance metrics across all runs for each scenario and size. Nurse Ratios uses a "Stressed Staffing" baseline, while Fast Track and Split-Flow use ("High Volume") as motivated in the Methods section. Observing the 95\% confidence intervals for the key metrics across all runs, we find that Split-Flow (red colors) consistently shows shorter LoS/wait times; an effect that is particularly pronounced in X-large configurations. In contrast, Fast Track (blue colors) provides less improvement in the X-large ED, while Nurse Ratio limiting has no effect as expected.

We performed independent t-tests (Welch's t-test) and calculated Cohen's d effect sizes comparing each intervention to the Baseline of its respective size (Table \ref{tab:stat_significance}). Many of the observed improvements are statistically significant in our simulation. The expected exception is Nurse Ratios, which shows no significant effect on either LoS or Wait Times across all sizes.

\begin{table*}[tb]
\centering
\begin{tabular}{llccc}
\toprule
Intervention & Metric & Medium & Large & X-large \\
\midrule
\textbf{Fast Track} & LoS & $d=3.17, p<0.001$ & $d=3.88, p<0.001$ & $d=1.53, p<0.001$ \\
 & Wait Time & $d=1.66, p<0.001$ & $d=2.64, p<0.001$ & NS ($p=0.31$) \\
 & LWBS (\%) & $d=4.36, p<0.001$ & $d=7.66, p<0.001$ & $d=7.95, p<0.001$ \\
 & Mortality (\%) & $d=1.28, p<0.001$ & $d=2.04, p<0.001$ & NS ($p=0.28$) \\
\midrule
\textbf{Nurse Ratios} & LoS & NS ($p=0.91$) & NS ($p=0.66$) & $d=-0.86, p=0.002$ \\
 & Wait Time & NS ($p=0.58$) & NS ($p=0.77$) & $d=-0.75, p=0.005$ \\
 & LWBS (\%) & NS ($p=0.17$) & $d=1.31, p<0.001$ & $d=2.44, p<0.001$ \\
 & Mortality (\%) & NS ($p=0.35$) & NS ($p=0.28$) & $d=-0.82, p=0.002$ \\
\midrule
\textbf{Split-Flow} & LoS & $d=2.89, p<0.001$ & $d=4.11, p<0.001$ & $d=8.88, p<0.001$ \\
 & Wait Time & $d=2.50, p<0.001$ & $d=5.01, p<0.001$ & $d=8.25, p<0.001$ \\
 & LWBS (\%) & NS ($p=0.53$) & NS ($p=0.98$) & $d=-3.78, p<0.001$ \\
 & Mortality (\%) & $d=0.66, p=0.013$ & $d=0.87, p=0.001$ & $d=2.38, p<0.001$ \\
\bottomrule
\end{tabular}
\caption{Detailed Statistical Significance Analysis (Cohen's $d$ and $p$-values) across ED Sizes.}
\label{tab:stat_significance}
\end{table*}

\begin{figure*}[t]
    \centering
    \includegraphics[width=.85\linewidth]{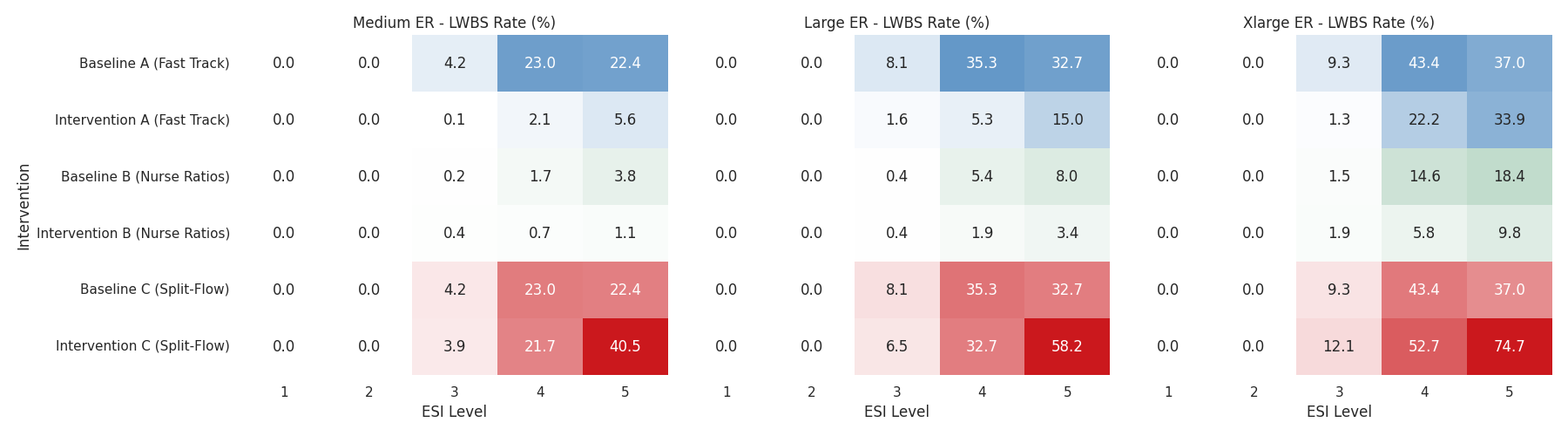}
    \caption{Left Without Being Seen (LWBS) Heatmap bsy ED Size and Acuity (true ESI levels)}
    \label{fig:lwbs_heatmap}
\end{figure*}

Patient retention is a key indicator of process efficiency, measured as Left Without Being Seen (LWBS) rate. The heatmaps in Figure \ref{fig:lwbs_heatmap} show LWBS rates broken down by ESI level for each size, again color-coded with two consecutive rows depicting one intervention. This paints a mixed picture of intervention effectiveness. While Nurse Ratio shows a slight, but consistent reduction in LWBS across all sizes, Fast Track increases LWBS in all settings. This might be a consequence from resource reallocation away from the main ED flow. Split-Flow, however, demonstrates a reduction in LWBS rates in Medium and Large environments, yet an increase in X-large settings which might be explained by capacity constraints or operational complexities unique to very large EDs.

\begin{figure*}[t]
    \centering
    \includegraphics[width=.85\linewidth]{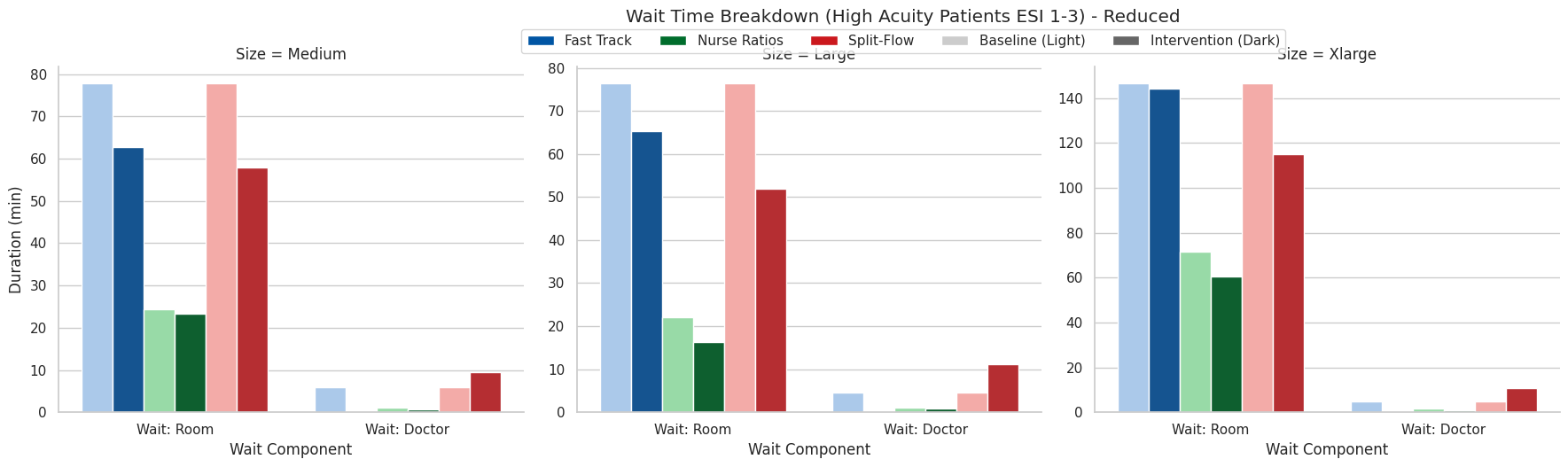}
    \caption{Wait Time Breakdown for High Acuity Patients (true ESI levels 1-2); wait times for triage and nurse omitted (insignificant or zero)}
    \label{fig:high_acuity_wait}
\end{figure*}

Lastly, Figure  \ref{fig:high_acuity_wait} shows the High Acuity Wait Time Breakdown. The time ESI 1-2 patients wait for treatment is a critical safety metric. Concordant with prior literature, Split-Flow significantly reduces wait times for high-acuity patients, while Fast Track and Nurse Ratios have less pronounced effects. All interventions help reduce the tail of the wait time distribution, which is crucial for patient outcomes in emergency settings. It is also visible that in the simulation wait times are predominantly driven by unavailability of treatment rooms rather than provider availability, highlighting the importance of spatial resource management in ED settings.

\subsection{Validation Study Qualitative Results}

The Fast Track intervention, as discussed by \cite{Berger-vonOrelli2025} and \cite{Bonalumi2017}, was strongly validated for its impact on Length of Stay (LoS). It segregated low-acuity patients from the regular flow, freeing resources for those in need, and the simulation showed a reduction in LoS of about 6-27\% (expected: 20-30\%). Its impact on overall wait time was equally pronounced and statistically significant (24 and 26\% for medium and large size), but not for X-large where saturation is observed. Fast Track reduced LWBS in all three settings dramatically (between 50 and 90\%) which is in accordance with improvements reported in literature.

Regarding nurse-to-patient ratios, modeled after California Assembly Bill 394 and the work of \cite{spetz2004california}, the results indicated a neutral impact on throughput across all sizes of EDs. Implementing a 1:4 ratio did not significantly degrade LoS or Wait Times, nor did it provide a speedup by improving on staff fatigue. In the X-large size ED, a small, but significant increase in those metrics had to be reported. LWBS, however, was improved for all size by about 50\%. This suggests that at the tested volumes, nurses did not reach the fatigue thresholds where ratio constraints would become a dominant factor in patient flow. It also implies that the waiting room can be the next bottleneck as nurses once their ratio is met can't take on more patients. These results are consistent with the report of \cite{doi:10.1377/hlthaff.26.3.853}, who pointed out that mandated ratios do improve on nurse satisfaction and patient care quality, but may show limited effects on throughput metrics.

Split-Flow emerged as the most effective intervention, consistent with findings by \cite{rowe2011use}. By transforming the patient care pathway dynamically for patients in need by having a doctor see patients already in the triage area, this approach reduced LoS by about 25\% and Wait Time by between 33 and 42\% (expected: 20-35\%). It affected LWBS negatively for X-large EDs since the Physician at Triage becomes the new bottleneck making patients choose to leave. Crucially, however, outcomes improve for those who stay: the significant reduction in fatalities effected by this intervention suggests that "Door to Provider" time is a critical safety driver.

\subsection{Multi-Agent System Qualitative Results}

In overall 15 experimental runs on three configurations (baseline, high-load, low-staffing) of the Large ED setting, we were able to demonstrate the full functionality of the Multi-Agent System (MAS). The specialist agents proposed room and staff additions or removals based on observed staff bottlenecks or long wait times for resources in the simulated ED. The orchestrator agent successfully coordinated these suggestions, evaluating their potential impact on overall ED performance before creating action messages to the simulation. Concurrently, the MAS wrote experiences to memory. The system also writes extensive log files to capture the decision-making process and outcomes of these interventions (compare Appendix for plots and details). Especially targeting LLM finetuning or reinforcement learning strategies, we store not only the path after the MAS intervention, but also the baseline path without. This can be the basis for deeper analysis and refinement of the MAS strategies, or serve as training data for machine learning models to enhance agent decision-making in subsequent iterations.


\section{Conclusions and Further Work}

The simulation of an ED based on Discrete Event and Agent-Based Modeling is able to simulate not only complex pathways based on the different diseases, but also includes realistic parameters like fatiguing staff, slowdown, and medical errors etc. as behaviors of agents emerging over time. We were able to show with our validation approach that the model accurately reproduces real-world dynamics and can be used to test various resource optimization strategies. Together, our findings align well with existing literature, reinforcing the validity of the model and its assumptions.

A model like this is not meant to explain real-world dynamics in full detail, but rather to provide a sandbox for testing various interventions and strategies before implementing them in practice. Since the simulation allows to log all events over time, it lends itself to generate training material for machine-learning based decision support systems as well as to optimize resource allocation in real-time based on the current state of the ED. 

We were able to integrate a working version of such an optimization employing a Multi-Agent System based on a locally running LLM. While this MAS was not in the scope of this validation study, it showed promising initial results and will be the subject of future work. It is equipped with a memory architecture enabling it to derive episodic knowledge from prior runs and use this to optimize resource allocation and patient flow in the ED Digital Twin. With this in hand, we will in close collaboration with clinical partners evaluate the system in practice by modeling a real-world ED in size, spatial configuration, and patient demographics to validate the power of the system and explore the feasibility of autonomous agents for a real-world setting further.

\appendix

\section*{Ethical Statement}

There are no ethical issues. All data used in this work is fully simulated. No human or animal subjects were involved in this research, and no sensitive or personal data was used.

\section*{Acknowledgments and AI Use Declaration}

This contribution has been collaborative work together with colleagues from the Asklepios ProResearch group, the Asklepios Department Heads of three Emergency departments in Hamburg, Germany, and several colleagues from the Fraunhofer Institute for Digital Medicine MEVIS in Bremen, Germany. We would like to thank all of them for their support and fruitful discussions.

LLMs have been used to implement the DES-ABM simulation framework, including data analysis tools and summarizing result reports. In this process, human intervention has been key to the development, steering the implementation, defining the design of the framework, and constructing metrics and visualizations. LLMs also aided to format specific parts of the manuscript, like tables. All methods and results summaries provided by LLMs have undergone a complete rewrite by the authors. All other parts of this contribution have been written by the authors; in particular all references have been researched and compiled manually.

\bibliographystyle{named}
\bibliography{ed_sim}

\section*{Additional Material}
\label{chap:appendix}

\subsection*{Quantitative Results}

\begin{table*}[t]
\centering
\begin{tabular}{lllll}
\toprule
Size & Scenario & Avg LOS (min) & Avg Wait Time (min) & LWBS (\%) \\
\midrule
\textbf{Medium} & \textit{Baseline (High Vol)} & \textit{208.7} & \textit{116.7} & \textit{9.6\%} \\
 & Fast Track & 152.3 (-27.0\%) & 88.6 (-24.1\%) & 1.0\% (-89.9\%) \\
 & Split-Flow & 157.6 (-24.5\%) & 78.3 (-32.9\%) & 10.1\% (+5.3\%) \\
 \cmidrule{2-5}
 & \textit{Baseline (Stressed)} & \textit{104.5} & \textit{34.6} & \textit{0.8\%} \\
 & Nurse Ratios & 105.0 (+0.5\%) & 36.4 (+5.2\%) & 0.4\% (-45.6\%) \\
\midrule
\textbf{Large} & \textit{Baseline (High Vol)} & \textit{212.5} & \textit{122.8} & \textit{15.4\%} \\
 & Fast Track & 161.2 (-24.1\%) & 90.5 (-26.3\%) & 3.1\% (-79.7\%) \\
 & Split-Flow & 161.8 (-23.8\%) & 71.9 (-41.5\%) & 15.4\% (-0.1\%) \\
 \cmidrule{2-5}
 & \textit{Baseline (Stressed)} & \textit{116.1} & \textit{37.5} & \textit{2.2\%} \\
 & Nurse Ratios & 117.6 (+1.3\%) & 38.3 (+2.1\%) & 0.9\% (-57.6\%) \\
\midrule
\textbf{X-large} & \textit{Baseline (High Vol)} & \textit{291.7} & \textit{191.9} & \textit{19.0\%} \\
 & Fast Track & 273.4 (-6.3\%) & 189.0 (-1.5\%) & 9.4\% (-50.4\%) \\
 & Split-Flow & 211.9 (-27.4\%) & 124.5 (-35.1\%) & 24.5\% (+29.3\%) \\
 \cmidrule{2-5}
 & \textit{Baseline (Stressed)} & \textit{193.6} & \textit{103.7} & \textit{5.9\%} \\
 & Nurse Ratios & 203.7 (+5.2\%) & 112.2 (+8.1\%) & 3.1\% (-47.8\%) \\
\bottomrule
\end{tabular}
\caption{Quantitative Results Comparison across ED Sizes}
\label{tab:quant_results_all}
\end{table*}

Table~\ref{tab:quant_results_all} presents a comparison of quantitative results across different emergency department (ED) sizes (Medium, Large, and X-large). The table details the average length of stay (LoS), average wait time, and the proportion of patients who left without being seen (LWBS) for the baselines and interventions. The High Volume baseline is contrastet with the Fast Track and Split-Flow interventions, and the Stressed baseline condition with Nurse Ratios adjustments. Percentage changes from the baseline are also provided for each scenario to highlight performance improvements or declines.

\subsection*{Additional Plots}

\begin{figure*}[t]
    \centering
    \includegraphics[width=.8\linewidth]{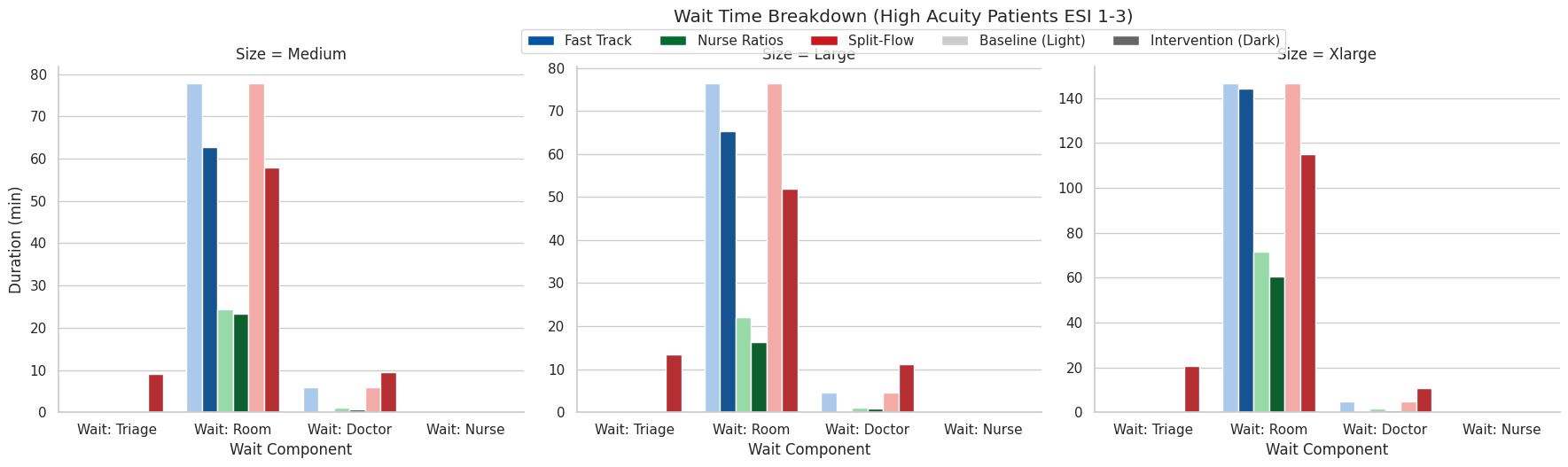}
    \caption{Wait time breakdown for high acuity patients (ESI levels 1 and 2)}
    \label{fig:high_acuity_wait_full}
\end{figure*}

\begin{figure*}[t]
    \centering
    \includegraphics[width=.8\linewidth]{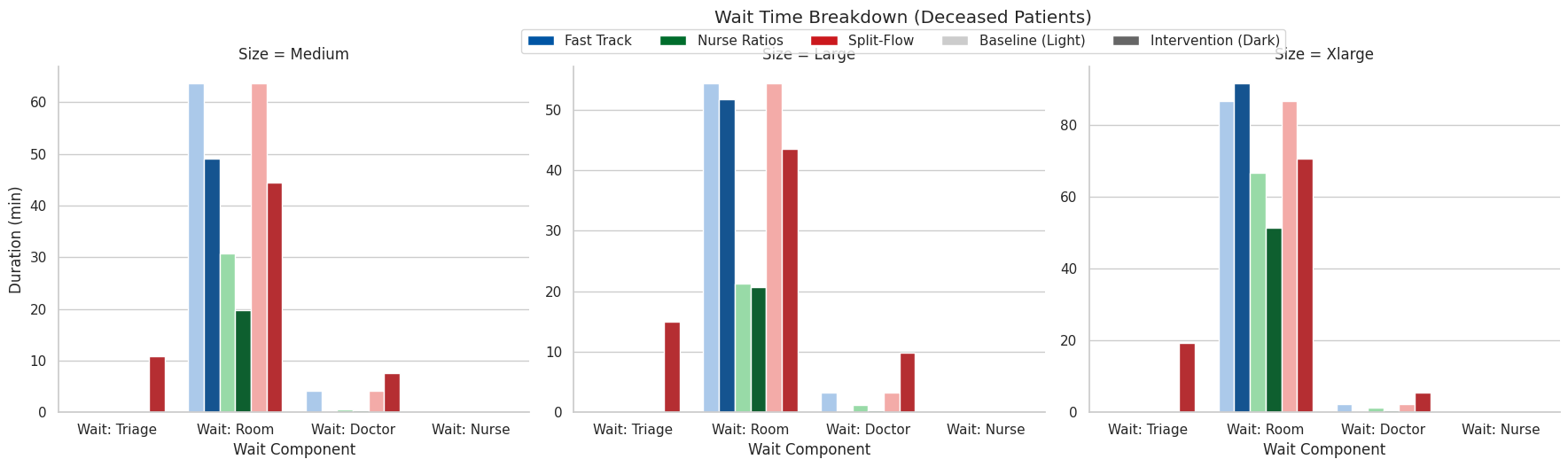}
    \caption{Wait time breakdown for deceased patients}
    \label{fig:deceased_wait_full}
\end{figure*}

Figure~\ref{fig:high_acuity_wait_full} illustrates the wait time breakdown for high acuity patients without ommitting any wait times. It shows that some patients wait for triage only in the X-large ED setting which generally is loaded slighly more than the other two sizes. Figure~\ref{fig:deceased_wait_full} illustrates the wait time breakdown for deceased patients without ommitting any wait times.

\begin{figure*}[t]
    \centering
    \includegraphics[width=.8\linewidth]{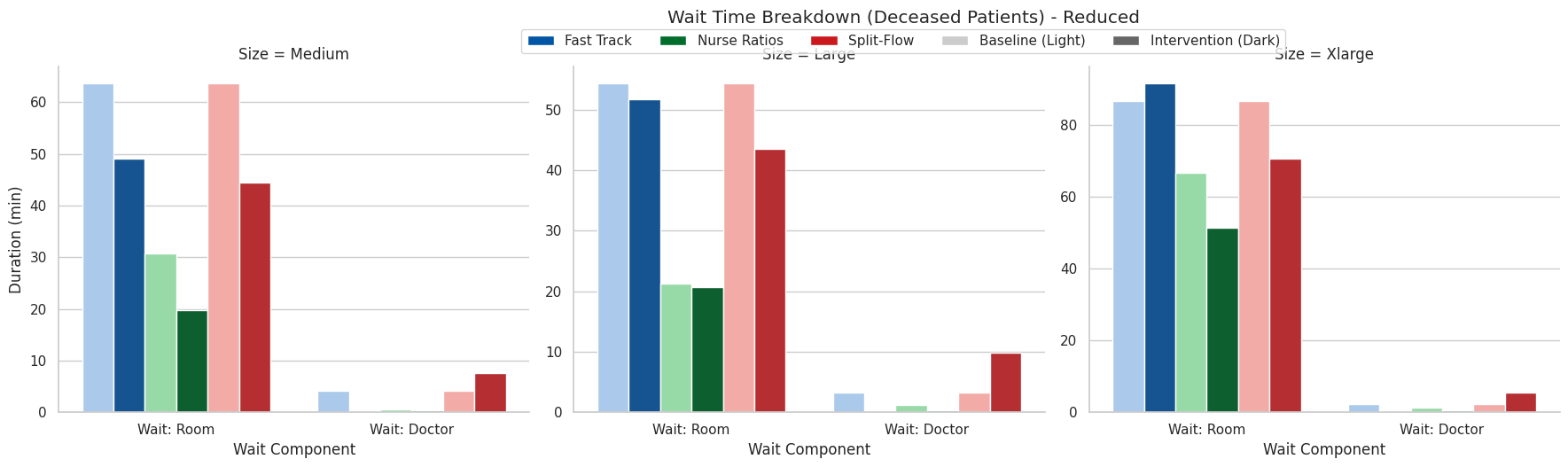}
    \caption{Wait time breakdown for deceased patients; wait times for triage and nurse omitted}
    \label{fig:deceased_wait}
\end{figure*}

Figure~\ref{fig:deceased_wait} illustrates the wait time breakdown for deceased patients, this time excluding wait times for triage and nurse assessments.

\begin{figure*}[t]
    \centering
    \includegraphics[width=\linewidth]{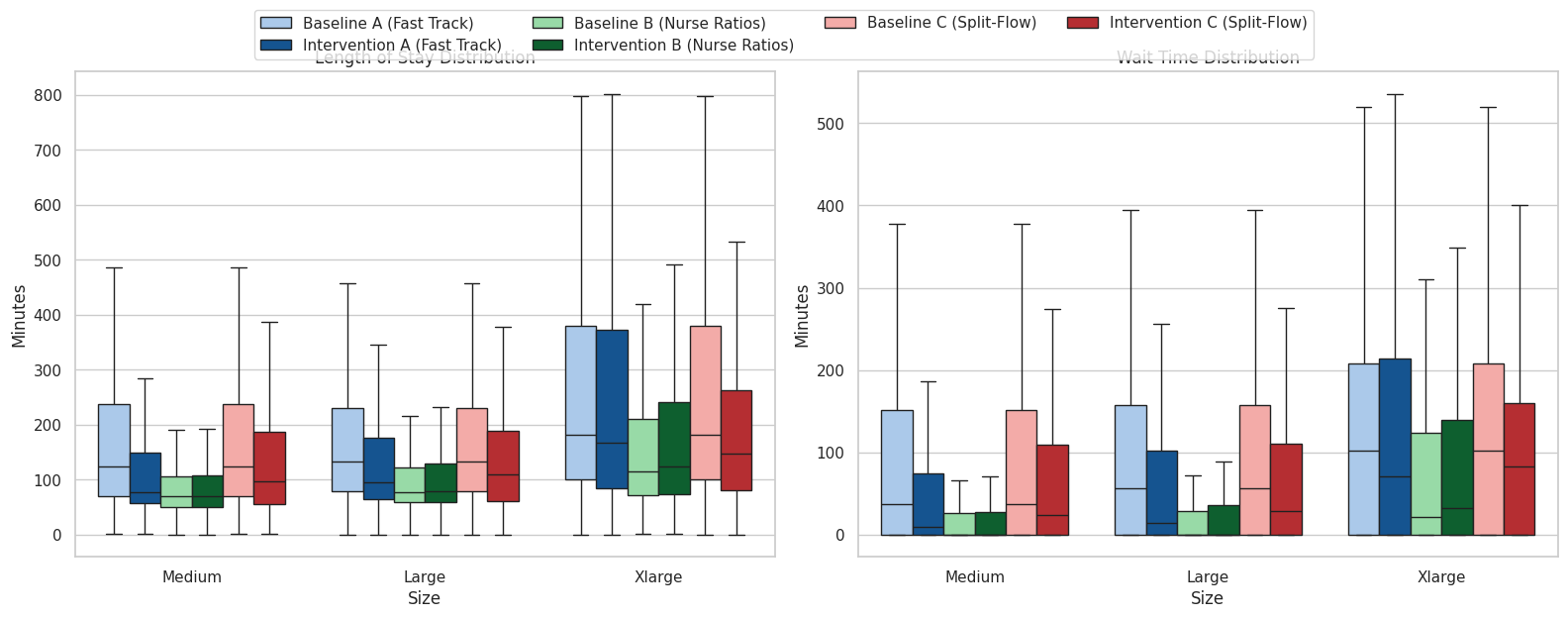}
    \caption{Analysis of performance indicator distributions across ED sizes and interventions. Like colors (blue, green, red) represent the interventions within each size category, and their shades baseline vs. intervention outcome.}
    \label{fig:unified_dist}
\end{figure*}

Figure~\ref{fig:unified_dist} provides an analysis of performance indicator distributions across different emergency department (ED) sizes and interventions. The plot uses color coding to represent various interventions within each size category, with shades indicating baseline versus intervention outcomes. This visualization aids in understanding how different ED sizes respond to specific interventions in terms of key performance metrics.

\subsection*{Temporal Evolution of KPIs}

\begin{figure*}[t]
    \centering
    \includegraphics[width=.8\linewidth]{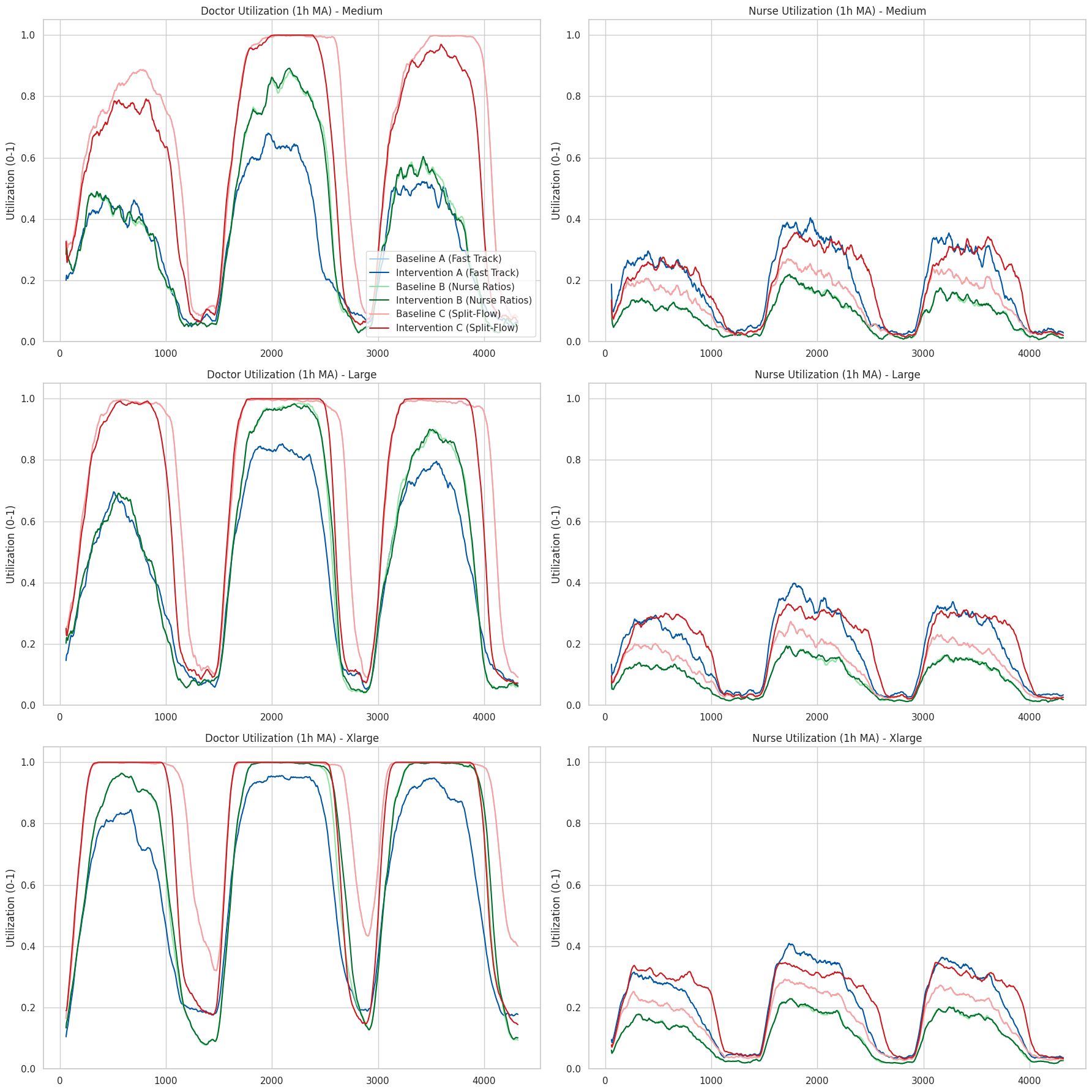}
    \caption{Wait time breakdown for deceased patients; wait times for triage and nurse omitted}
    \label{fig:temporal_staff}
\end{figure*}

\begin{figure*}[t]
    \centering
    \includegraphics[width=.8\linewidth]{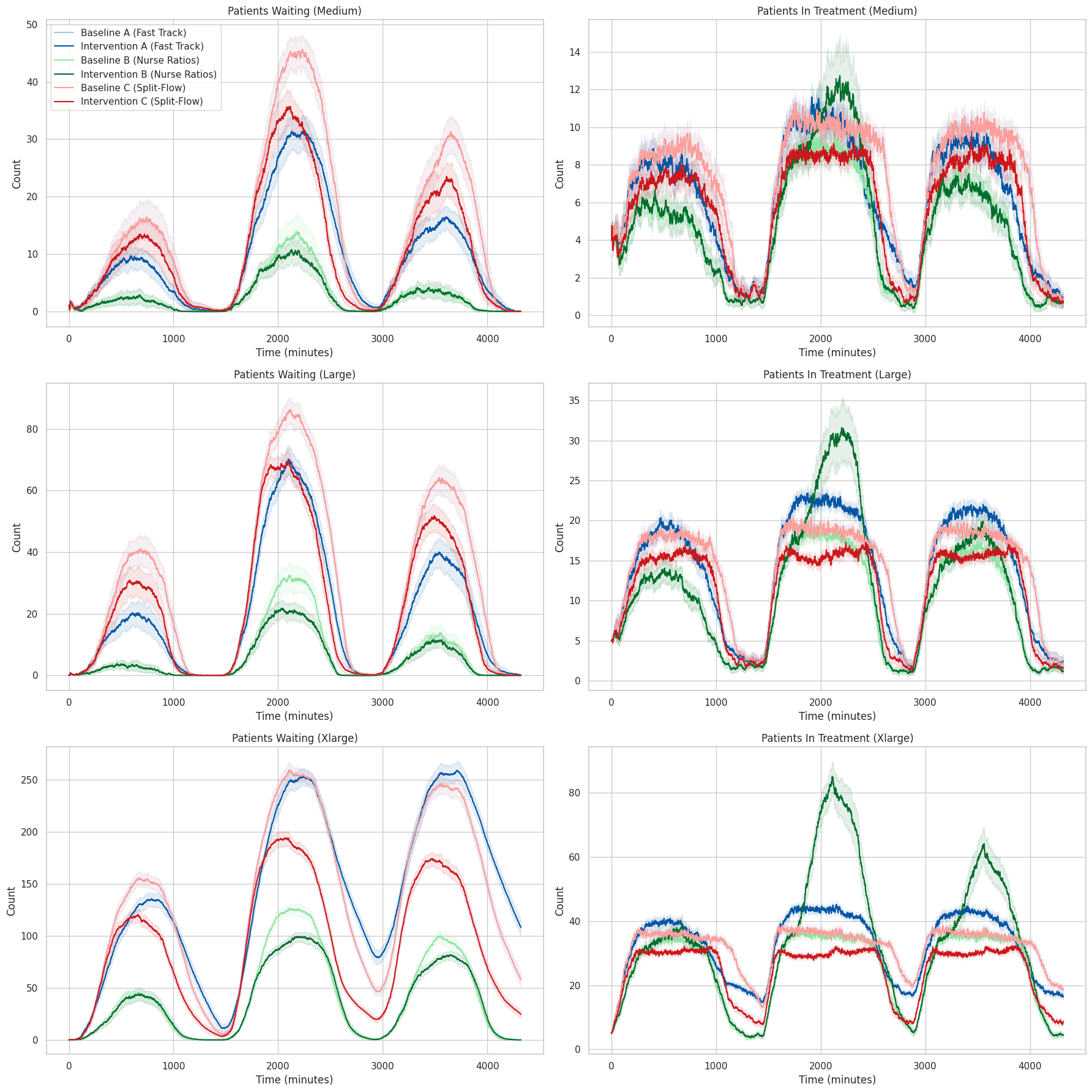}
    \caption{Wait time breakdown for deceased patients; wait times for triage and nurse omitted}
    \label{fig:temporal_pat_stat}
\end{figure*}

In two plots (Figures \ref{fig:temporal_staff} and \ref{fig:temporal_pat_stat}) we show the evolution of staff utilization and patient status over the three simulated days to give an intuition about the periodicity implemented in the ED simulation. The color coding with lighter color shades for the baselines and darker shades for the post-intervention situation are the same as in all other plots. The shaded areas indicate the variability over the 30 simulation runs per scenario.

\subsection*{Preliminary MAS Results}

To establish a first baseline comparing the Multi-Agent System (MAS) proposed interventions with the three scientifically esttablished ones (Nurse Ratios, Fast Track, Split-Flow), we ran three experiments for the Large ED size: The default configuration, a High-Load configuration similar to the one used for Fast-Track and Split-Flow, and a Low-Staff configuration as in the Nurse Ratios intervention baseline. Each experiment was run for 5 iterations with different random seeds. 

Our preliminary MAS implementation uses specialist agents looking at specific load-related, staffing-related and resource-related questions; in addition, a data analyst globally observes the ED metrics evolution. All of them independently post their evaluation of the batch of 60 steps (equalling 60 simulated minutes) to the messaging system, where the orchestrator agent listens until all specialists have posted their requests (or a "Silent" signal). The MAS orchestrator then calls a tool interface of the simulation to perform the harmonized actions. The MAS was for these experiments only allowed to open or close rooms of different types, and to allocate or dismiss any type of staff. Also it has only limited perception of the ED state, both in terms of parameters and time horizon. 

\begin{figure*}[t]
    \centering
    \includegraphics[width=.8\linewidth]{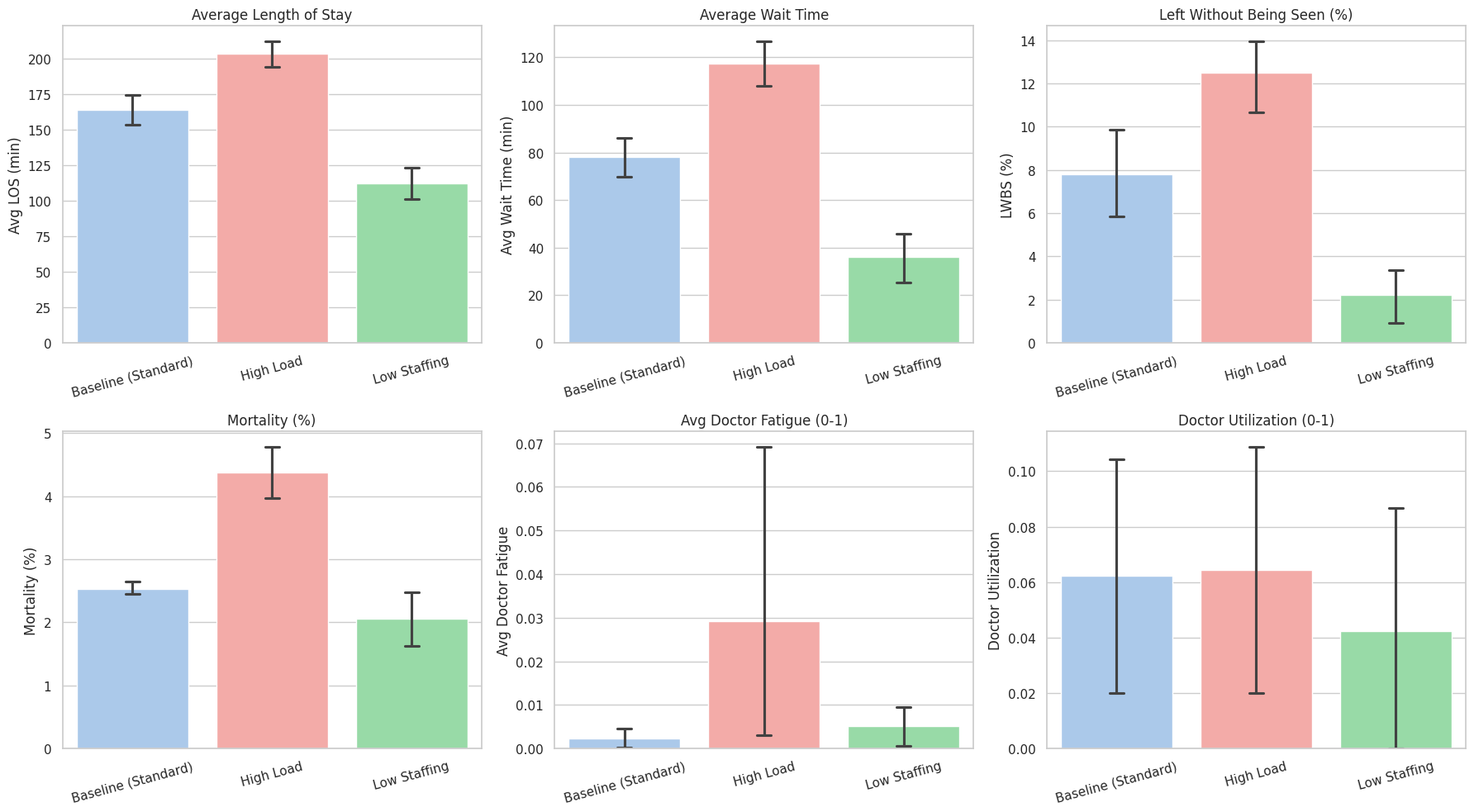}
    \caption{Wait time breakdown for deceased patients; wait times for triage and nurse omitted}
    \label{fig:temporal_metrics_mas}
\end{figure*}

The plot in Figure~\ref{fig:temporal_metrics_mas} shows the metrics comparison between the MAS runs for the Large ED. These can be compared to the corresponding Large ED results in Figure~\ref{fig:unified_metrics}. The metrics marginally improved over the baselines; statistical significance has not been tested yet. 

\begin{figure*}[t]
    \centering
    \includegraphics[width=.8\linewidth]{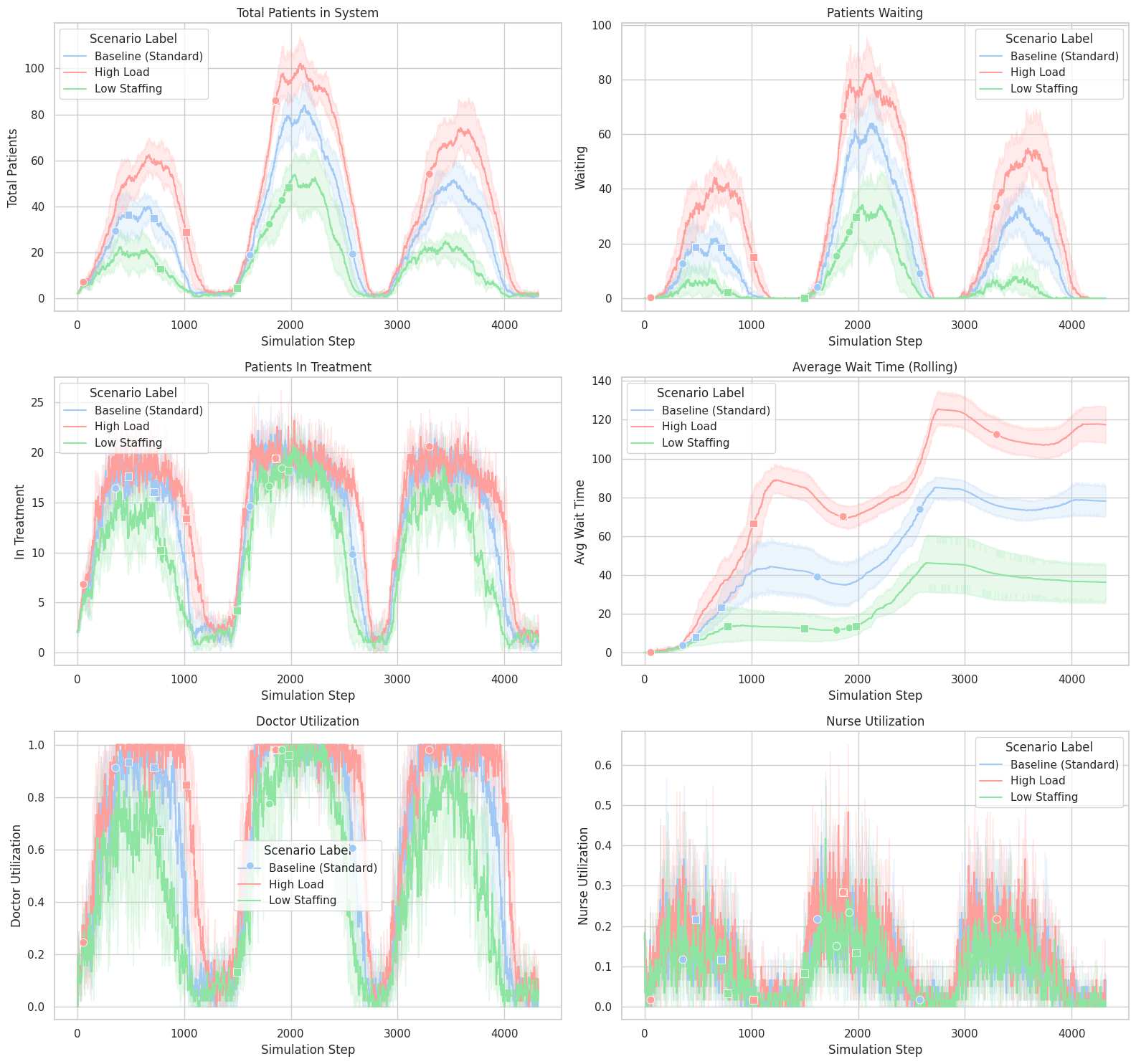}
    \caption{Wait time breakdown for deceased patients; wait times for triage and nurse omitted}
    \label{fig:temporal_comparison_mas}
\end{figure*}

To understand the impact of the MAS interventions over time, Figure~\ref{fig:temporal_comparison_mas} shows the temporal evolution of several metrics for the three MAS experiments. In the plots, dots represent the staff-related interventions, squares the rooms-related ones. In these analyses, there is no clear strategy emerging yet. Reasons for this are manifold. Most importantly, while in these runs, the MAS was allowed to write experiences to memory, the memory was initially empty, so that the agents started with no established knowledge. Also, neither have we conducted any prompt engineering, nor any experimentation into short-term and long-term memory management or variants of temporal horizons the MAS is allowed to read. These improvements are currently ongoing.

\subsection*{Floor Plans}

\begin{figure*}[tb]
    \centering
        \includegraphics[width=\linewidth]{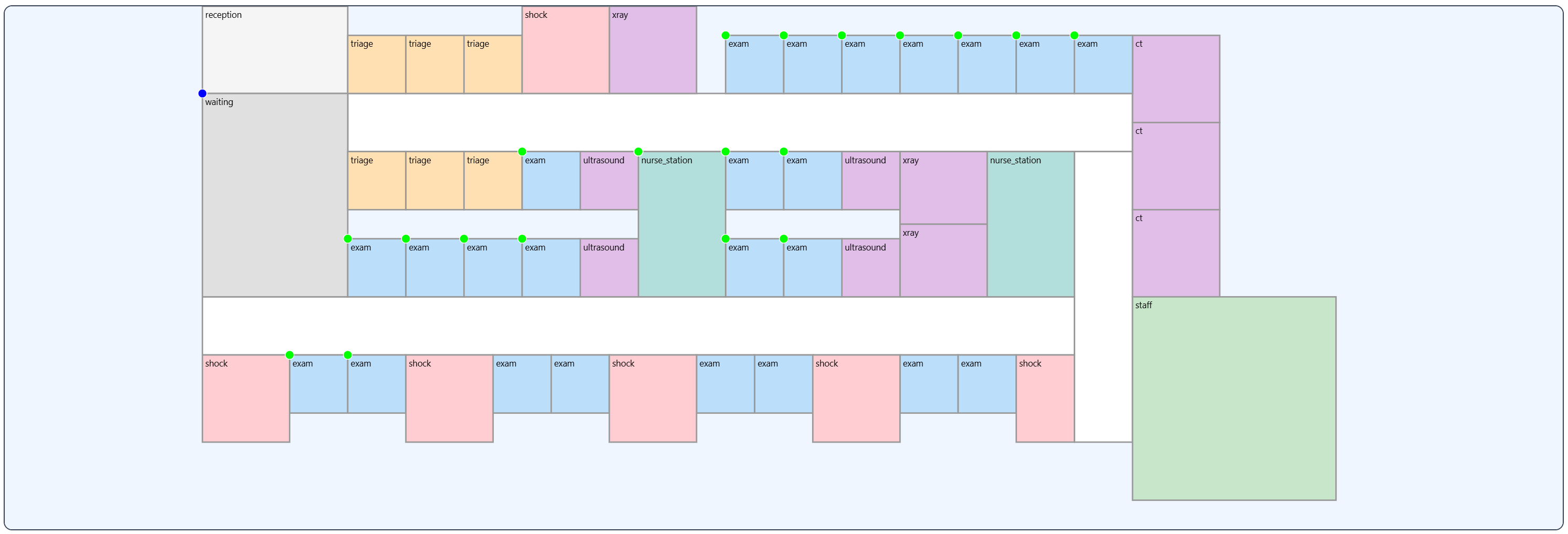}
    \caption{Exemplary floor plan for X-large ED size used in the simulation study.}
    \label{fig:floorplans}
\end{figure*}

The floor plans used in this study have been suggested by the coding agent based on typical ED layouts and recommendations found in literature. For the statistical analyses, they have been scaled to four sizes (small, medium, large, extra-large), out of which only the largest three have been used, while the smallest size served as a functional test scenario during model development. Staffing and room counts were kept as suggested by the model. The LLM-provided floor plans were usually implausible and were redrafted by hand to obtain more realistic layouts; shown in Figure~\ref{fig:floorplans} for the largest size. Colored dots indicate patients and staff assigned to areas and rooms. 

The layout provides a visual representation of the spatial organization within the ED, which in reality is a critical factor influencing patient flow and overall operational efficiency. The currently used floor plans are not modeled after any specific real-world ED. Also, note that with the step size setting of one minute, actual distances in the floor plans will usually be traversed within one or very few simulation step, so that no path planning is currently implemented. Agents instead have a traveling speed that determines the number of steps they need to arrive at their destination. Their speed also deteriorates when they move equipment or are fatiguing. Floors also can not be congested in the current model, but all room types have maximum occupancy parameters that could limit floor usage. 

\end{document}